# Adaptive Sensing for Learning Nonstationary Environment Models

Sahil Garg, Amarjeet Singh, Fabio Ramos
sahilgar@usc.edu, amarjeet@iiitd.ac.in, fabio.ramos@sydney.edu.au

*Abstract*— Most environmental phenomena, such as wind profiles, ozone concentration and sunlight distribution under a forest canopy, exhibit nonstationary dynamics i.e. *phenomenon variation* change depending on the location and time of occurrence. Non-stationary dynamics pose both theoretical and practical challenges to statistical machine learning algorithms aiming to accurately capture the complexities governing the evolution of such processes. In this paper, we address the sampling aspects of the problem of learning nonstationary spatio-temporal models, and propose an efficient yet simple algorithm - *LISAL*. The core idea in *LISAL* is to learn two models using *Gaussian processes (GPs)* wherein the first is a nonstationary GP directly modeling the phenomenon. The second model uses a stationary GP representing a latent space corresponding to changes in dynamics, or the nonstationarity characteristics of the first model. *LISAL* involves adaptively sampling the latent space dynamics using information theory quantities to reduce the computational cost during the learning phase. The relevance of *LISAL* is extensively validated using multiple real world datasets.

## I. INTRODUCTION

Understanding the spatio-temporal dynamics of environmental phenomena is of primal importance for sustainable development. From renewable energy generation problems such as wind speed estimation [1], to assessment of pollution impact such as ozone concentration in low altitudes [2], accurate and reliable spatio-temporal prediction is crucial from several aspects, including the development of new policies.

Given the complex dynamics of environmental phenomena (see Figure 1 for illustration), nonparametric Bayesian models have been the method of choice. These models are very flexible in capturing different characteristics of the phenomenon using different kernels, and naturally account for uncertainty in the predictions and noise in the observations. Amongst the nonparametric Bayesian methods, *Gaussian Processes* (GP) have been very popular in geostatistics and environmental sciences due to the analytical tractability for the posterior and marginal likelihood estimations [3], [4]. A GP is not only resilient to overfitting but also provides confidence levels which can be used to evaluate information metrics used for informative sensing such as *entropy*, *mutual information*, or the reduction of predictive variance [5], [6], [7].

The key challenge for robust prediction in environmental monitoring is to deal with varying parameterizations over the input domain necessary to model changes in the *dynamics of the phenomenon*, i.e. nonstationarity. To this end, nonstationary GP models have been proposed [8], [9], [10], [11], [12]. Of particular relevance for this work is the class of the generic nonstationary GPs (termed as N-GPs) that employs *local hyper-parameters* $\mathbf{z_V}$ across the entire input space $X_V \in \mathbb{R}^{n \times 3}$ to correspondingly represent the *latent space dynamics*. This work, although applicable to all N-GPs, focuses primarily on two nonstationary models: 1) *Process Convolution with Local Smoothing Kernels*-PCLSK [10]; and 2) *Latent Extension of Input Space*-LEIS [12]. PCLSK and LEIS models are chosen for their intuitiveness and flexibility in modeling the environmental phenomena as compared to alternative N-GPs such as Mixture of GPs [11].

Motivated by the work of [13], [14], we propose to employ a stationary GP (*latent GP*) for modeling the dynamics of local hyper-parameters of a N-GP model. The main contribution is to *adaptively sense the latent space dynamics using* LISAL- an algorithm that builds accurate representation of the latent dynamics by adaptively learning local hyper-parameters of the N-GP, and the latent GP. *LISAL* reduces the overall computational cost for learning the N-GPs while providing significant improvement in phenomenon modeling. Extensive empirical validation of *LISAL* using three environmental sensing datasets (see complex dynamics for the datasets in Figure 1(a), 1(b), 1(c)) demonstrate the relevance of the proposed work.

## II. GAUSSIAN PROCESSES

Gaussian Processes (GP) place a multivariate Gaussian distribution over the space of functions, $f(\mathbf{x})$, mapping the input space ($\mathbf{x} \in \mathbb{R}^p$) to output space: $f(\mathbf{x}) \sim \mathcal{N}(m(\mathbf{x}), K(\mathbf{x}, \mathbf{x}'))$, where, $m(\mathbf{x})$ is a mean function and $K(\mathbf{x}, \mathbf{x}')$ is a positive definite covariance function with hyper-parameters $\boldsymbol{\theta}$ [4]. Observations are modeled as noisy measurements of the function, $y = f(\mathbf{x}) + \epsilon$, where $\epsilon \sim \mathcal{N}(0, \sigma_n^2)$.

In a Bayesian setup, the hyper-parameters, $\boldsymbol{\theta}$, can be learned from training data, $\{X_V, \mathbf{y_V}\}$, by maximizing the *log marginal likelihood* ($lml$) defined as:

$$-\frac{1}{2}\mathbf{y_V^T} K_y^{-1} \mathbf{y_V} - \frac{1}{2}\log|K_y| - \frac{n}{2}\log 2\pi; K_y = K(X_V, X_V) + \sigma_n^2 I_n$$

Using the learned GP model, the joint Gaussian distribution of observed input locations $X_V \in \mathbb{R}^{n \times p}$ and test input locations $X_* \in \mathbb{R}^{n_* \times p}$, with zero mean function $m(.)$, can be written as:

$$\begin{bmatrix} \mathbf{y}_V \\ \mathbf{f}_* \end{bmatrix} \sim \mathcal{N}\left(0, \begin{bmatrix} K(X_V, X_V) + \sigma_n^2 I_n & K(X_V, X_*) \\ K(X_*, X_V) & K(X_*, X_*) \end{bmatrix}\right)$$

A popular covariance function between two input locations $\mathbf{x}, \mathbf{x}' \in \mathbb{R}^3$ is the squared exponential defined as:

$$K(\mathbf{x}, \mathbf{x}') = \sigma_f^2 \exp(-\frac{1}{2}(\mathbf{x} - \mathbf{x}')^T \boldsymbol{\Sigma}^{-1}(\mathbf{x} - \mathbf{x}')),$$

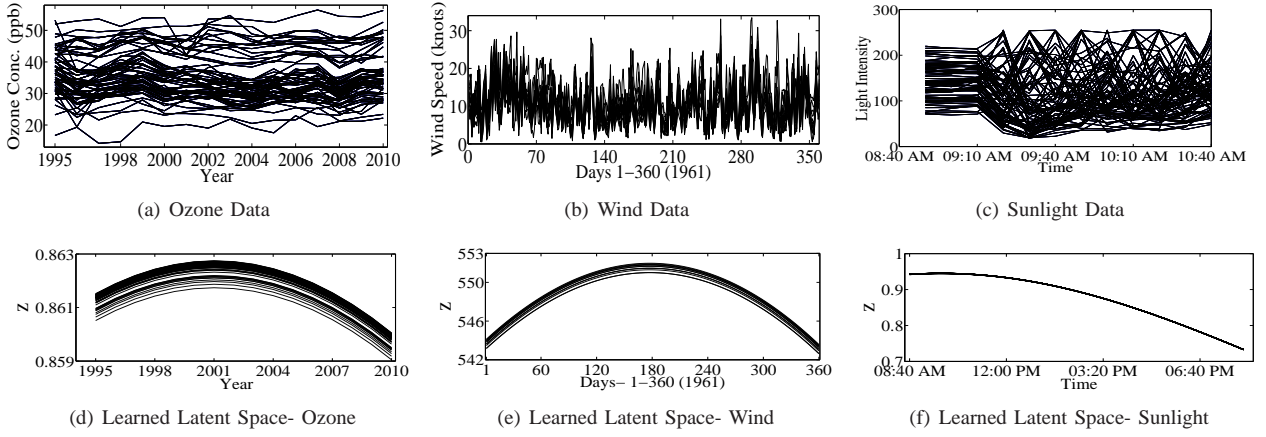

Fig. 1. Illustration of complex and diverse environmental dynamics that motivate this work. Each curve in 1(a), 1(b), 1(c) and 1(d), 1(e), 1(f), represents real and latent temporal dynamics respectively at a different spatial input location. 1(a) and 1(d) represent ozone conc. change and the corresponding latent dynamics for 60 locations. 1(b) and 1(e) illustrate wind speed change and the corresponding latent dynamics respectively for 12 locations. Sunlight intensity variation and the corresponding latent dynamics are shown in 1(c) and 1(f) respectively for 72 locations. Details of each dataset are discussed in Section V

where $\sigma_f$ is the signal variance, and $\Sigma$ is a matrix with length scale parameters. In the anisotropic case, with eigenvectors aligned along the coordinate axes, $\Sigma = diag([l_x^2, l_y^2, l_t^2])$, with $l_x, l_y, l_t \in \mathbb{R}$.

Next, we describe two nonstationary extensions of stationary Gaussian processes, denoted as N-GPs.

*1) Process Convolution with Local Smoothing Kernels- PCLSK:* Under the process convolution framework [10], [15], a nonstationary extension, $K_{ST}^{NS}$, of a stationary spatio-temporal covariance, $K_{ST}^{S}$, is proposed as [14]:

$$K_{ST}^{NS}(\mathbf{x_i}, \mathbf{x_j}) = \frac{|\Sigma_i|^{\frac{1}{4}}|\Sigma_j|^{\frac{1}{4}}}{|(\Sigma_i + \Sigma_j)/2|^{\frac{1}{2}}} . K_{ST}^{S}(\sqrt{q_{ij}^s}, \sqrt{q_{ij}^t}),$$

where $q_{ij}^s = (\mathbf{x}_{s_i} - \mathbf{x}_{s_j})^T \left(\frac{\Sigma_{i_s} + \Sigma_{j_s}}{2}\right)^{-1} (\mathbf{x}_{s_i} - \mathbf{x}_{s_j})$; $q_{ij}^t = (t_i - t_j)^2 \left(\frac{\Sigma_{i_t} + \Sigma_{j_t}}{2}\right)^{-1}$; $\mathbf{x} = [\mathbf{x}_s, t]$ (space and time coordinates); $\mathbf{x}_s \in \Re^2$ and $t \in \Re$; $\Sigma_i, \Sigma_j$ are local kernel matrices at $\mathbf{x_i}$ and $\mathbf{x_j}$ respectively. In anisotropic covariance functions, the local latent length scale parameters $\mathbf{l_{x_V}}, \mathbf{l_{y_V}}, \mathbf{l_{t_V}} \in \mathbb{R}^n$ across input space $X_V \in \mathbb{R}^{n \times 3}$ model the varying degree of phenomenon dynamics along the corresponding axes.

*2) Latent Extension of Input Space- LEIS:* In this approach, the real input space, $X_V$, is extended with latent space dimension, $\mathbf{l_V} \in \mathbb{R}^n$, resulting in a nonstationary covariance:

$$K_{ST}^{NS}(\mathbf{x}, \mathbf{x'}) = K_{ST}^{S} . \exp\left(-\frac{1}{2}\left(\frac{l - l'}{l_l}\right)^2\right),$$

where $l_l$ is a latent length scale parameter for the latent space. See [12] for details.

Based upon the discussion of *PCLSK*, *LEIS*, from now onwards, we can consider a general N-GP that employs global hyper-parameters $\theta_y$ and local hyper-parameters $\mathbf{z_V}$ (i.e. $\{\mathbf{l_{x_V}}, \mathbf{l_{y_V}}, \mathbf{l_{t_V}}\}$ for *PCLSK* and $\mathbf{l_V}$ for *LEIS*) across input space $X_V$. As an illustration, Figure 1(d), 1(e), 1(f) show $\mathbf{z_V}$ learned across $X_V$ for *LEIS* model using the proposed adaptive learning algorithm *LISAL*.

Local hyper-parameters $\mathbf{z_V}$ of N-GP corresponds to the latent space dynamics of an environmental phenomenon. Accuracy of $\mathbf{z_V}$ for representing the latent space dynamics depends upon intuitiveness of N-GP in modeling the specific phenomenon and accurate learning of $\mathbf{z_V}$. Considering the generic N-GPs, and specifics of *PCLSK*, *LEIS*, this work is entirely focused upon representative efficient learning of the latent space dynamics of a general class of environmental phenomena, as discussed in Section III and IV.

### III. LEARNING NONSTATIONARY GPS

Modeling environmental phenomena using N-GPs requires estimating local hyper-parameters $\mathbf{z_V}$ (across the input domain $X_V \in \mathbb{R}^{n \times 3}$), corresponding to the latent space dynamics, in addition to the global hyper-parameters $\theta_y$. Considering the typical large expanse of input space (high value of $n$), either of learning $\mathbf{z_V}$ by directly maximizing the marginal likelihood (see (1)) or sampling from the posterior (see (2)) is computationally expensive.

$$\{\mathbf{z_V^*}, \theta_y^*\} = argmax_{\mathbf{z_V}, \theta_y} p(\mathbf{y_V}|X_V, \mathbf{z_V}, \theta_y) \quad (1)$$

$$\{\mathbf{z_V^*}, \theta_y^*\} = argmax_{\mathbf{z_V}, \theta_y} p(\mathbf{z_V}, \theta_y|X_V, \mathbf{y_V}) \quad (2)$$

Therefore, in Section III-A, we introduce approximations from the existing literature that result in an efficient representation of the latent space dynamics. These approximations eventually motivate the need for adaptive learning algorithm *LISAL*, discussed in Section IV.

#### A. Latent Space Dynamics Representation

While the latent space dynamics are expected to be complex across space and time, the degree of nonstationarity of the latent dynamics is expected to be less than the real observable dynamics (see Figure 1 for comparison on the real and learned latent dynamics). Following existing work that uses this intuition [13], we employ a stationary GP with hyper-parameters $\boldsymbol{\theta_z}$ (termed as *latent GP* and represented as $GP_z$) for modeling the latent space dynamics. We then learn local hyper-parameters $\mathbf{z_M}$ across a sparse set of input locations $X_M \in \mathbb{R}^{m \times 3}$ s.t. $m << n$ (termed as *latent locations*) so as to infer joint Gaussian distribution $\mathbf{z_V}$ across $X_V$ by conditioning $\boldsymbol{\theta_z}$ on $\{X_M, \mathbf{z_M}\}$. It is important to note that *modeling the latent space dynamics with a stationary GP ensures that N-GP model does not result in overfitting with increase in the number of local hyper-parameters across the latent locations (i.e. increment on $m$)*, as is also apparent from our empirical results.

With the introduction of the *induced* locations $X_M$, expressions in (1) and (2) are transformed into optimization of intractable integrals as shown in (3) and (4) respectively.

$$\{\mathbf{z_M^*}, \boldsymbol{\theta_y^*}, \boldsymbol{\theta_z^*}\} = argmax_{\mathbf{z_M}, \boldsymbol{\theta_y}, \boldsymbol{\theta_z}} \int p(\mathbf{y_V}|X_V, \mathbf{z_V}, \boldsymbol{\theta_y}).$$
$$p(\mathbf{z_V}|X_V, X_M, \mathbf{z_M}, \boldsymbol{\theta_z}) \, d\mathbf{z_V} \quad (3)$$

$$\{\mathbf{z_M^*}, \boldsymbol{\theta_y^*}, \boldsymbol{\theta_z^*}\} = argmax_{\mathbf{z_M}, \boldsymbol{\theta_y}, \boldsymbol{\theta_z}} \int p(\mathbf{z_M}, \boldsymbol{\theta_z}|X_V, X_M, \mathbf{z_V}).p(\mathbf{z_V}, \boldsymbol{\theta_y}|X_V, \mathbf{y_V}) \, d\mathbf{z_V} \quad (4)$$

wherein the predictive joint Gaussian distribution $\mathbf{z_V}$ across $X_V$, is marginalized so as to learn local hyper-parameters ($\mathbf{z_M}$) across $X_M$ only.

[16] proposed *Gaussian approximation* that uses the law of iterated expectations and conditional variance, to analytically approximate the predictive joint Gaussian distribution in terms of predictive mean and predictive variance. To avoid complications of accounting for local hyper-parameters prediction uncertainty using predictive variance, it is proposed in [13], [14] to approximate the joint Gaussian distribution $\mathbf{z_V}$ with only predictive mean $\mathbf{z_V^m}$. So we apply the Gaussian approximation on the intractable integrals, shown in (3) and (4), by approximating $\mathbf{z_V}$ with $\mathbf{z_V^m}$, to create computationally tractable closed form expressions, as shown in (5) and (6) respectively.

$$\{\mathbf{z_M^*}, \boldsymbol{\theta_y^*}, \boldsymbol{\theta_z^*}\} = argmax_{\mathbf{z_M}, \boldsymbol{\theta_y}, \boldsymbol{\theta_z}} p(\mathbf{y_V}|X_V, \mathbf{z_V^m}, \boldsymbol{\theta_y}).$$
$$p(\mathbf{z_V^m}|X_V, X_M, \mathbf{z_M}, \boldsymbol{\theta_z}) \quad (5)$$

$$\{\mathbf{z_M^*}, \boldsymbol{\theta_y^*}, \boldsymbol{\theta_z^*}\} = argmax_{\mathbf{z_M}, \boldsymbol{\theta_y}, \boldsymbol{\theta_z}} p(\mathbf{z_M}, \boldsymbol{\theta_z}|X_V, X_M, \mathbf{z_V^m}).$$
$$p(\mathbf{z_V^m}, \boldsymbol{\theta_y}|X_V, \mathbf{y_V}) \quad (6)$$

wherein $\mathbf{z_V^m}$ is the predictive mean for local hyper-parameters, inferred by conditioning $\boldsymbol{\theta_z}$ on $\{X_M, \mathbf{z_M}\}$. $\mathbf{z_V^m}$ is employed as local hyper-parameters across $X_V$, in place of $\mathbf{z_V}$, to the N-GP, and corresponds to the latent space dynamics of the modeled environmental phenomenon.

Clearly, for $m << n$, computational cost for optimization in (5) and 6, will be significantly smaller than optimization in (1) and (2) respectively. However, we need to establish if the concept of induced latent locations, compromises on the accurate representation of latent space dynamics.

**Algorithm 1** LISAL

1: **Input:** $\{X_V \in \mathbb{R}^{n \times 3}, \mathbf{y_V} \in \mathbb{R}^n\}, m_1, m_2, c$
2: **Output:** $\boldsymbol{\theta_y^*}, \boldsymbol{\theta_z^*}, X_M^* \in \mathbb{R}^{m \times 3}, \mathbf{z_M^*} \in \mathbb{R}^m$
3: $\boldsymbol{\theta_{z_0}^*} = argmax_{\boldsymbol{\theta_{z_0}}} p(\mathbf{y_V}|X_V, \boldsymbol{\theta_{z_0}})$
4: $X_{M_1}^* = argmax_{X_{M_1}} \mathbb{I}(X_{M_1}; X_{V \setminus M_1}|\boldsymbol{\theta_{z_0}^*})$
5: $\{\boldsymbol{\theta_{y_1}^*}, \boldsymbol{\theta_{z_1}^*}, \mathbf{z_{M_1}^*}\} = snsLt(X_V, \mathbf{y_V}, X_{M_1}^*, \boldsymbol{\theta_{y_1}}, \boldsymbol{\theta_{z_1}}, \mathbf{z_{M_1}})$
6: **for** $i = 1 \to c$ **do**
7: $\quad X_{M_{i+1}}^* = argmax_{X_{M_{i+1}}} \mathbb{I}(\{X_{M_{i+1}}, X_{M_{1...i}}^*\}; X_{V \setminus M_{1...i+1}}|\boldsymbol{\theta_{z_i}^*}, X_{M_{1...i}}^*)$
8: $\quad \{\boldsymbol{\theta_{y_{i+1}}^*}, \boldsymbol{\theta_{z_{i+1}}^*}, \mathbf{z_{M_{i+1}}^*}\} = snsLt(X_V, \mathbf{y_V}, X_{M_{i+1}}^*, \boldsymbol{\theta_{y_{i+1}}}, \boldsymbol{\theta_{z_{i+1}}}, \mathbf{z_{M_{i+1}}}, X_{M_{0...i}}^*, \mathbf{z_{M_{0...i}}^*})$
9: **end for**

*B. Limitations of Induced Points Representation*

In reference to (5), if the selection of $X_M$ is not representative of $X_V$ in the context of latent dynamics, the predictive mean $\mathbf{z_V^m}$ (inferred by conditioning $\boldsymbol{\theta_z}$ on $\{X_M, \mathbf{z_M}\}$), will also inaccurately represent the latent dynamics across $X_V$. Inaccurate prediction of $\mathbf{z_V^m}$ eventually constrains the learning of $\mathbf{z_V^m}$ while marginalizing the real observations $\mathbf{y_V}$. Further, inaccurate learning of $\mathbf{z_V^m}$ leads to inaccurate learning of $\boldsymbol{\theta_z}$ and $\mathbf{z_M}$. Similar argument applies to optimization in (6) as well. How do we then select induced latent locations $X_M$ that are *informative* about latent dynamics across $X_V$, so as to perform accurate learning of N-GP?

*Mutual information* criterion has been a favorable choice amongst the other Bayesian criteria, such as *entropy* and *reduction of predictive variance*, for the problem of informative sensing of environmental dynamics [17], [18], [19]. In [14], it is proposed to quantify mutual information on real observations $\mathbf{y_V}$ (see (7)) for learning latent locations $X_M$.

$$X_M^* = argmax_{X_M: X_M \subset X_V} \mathbb{I}(X_M; X_{V \setminus M}|\boldsymbol{\theta_{z_0}^*});$$
$$\boldsymbol{\theta_{z_0}^*} = argmax_{\boldsymbol{\theta_{z_0}}} p(\mathbf{y_V}|X_V, \boldsymbol{\theta_{z_0}}) \quad (7)$$

However, it is hard to establish that *latent locations selected using mutual information on real observable dynamics will also be informative about the latent space dynamics*. Therefore, we propose *LISAL* for adaptive informative sensing of the latent space dynamics.

IV. LATENT INFORMATIVE SENSING WITH ADAPTIVE LEARNING- *LISAL*

*LISAL* incrementally decides on the informative subset of latent locations ($X_M \subset X_V$) using adaptive sensing of latent space dynamics.

We first learn a stationary GP model, with hyper-parameters $\boldsymbol{\theta_{z_0}}$, using the real observations (Line 3 in Algorithm 1). Similar to the overall approach followed in [14], *LISAL* then learns a preliminary set of latent locations $X_{M_1} \in \mathbb{R}^{m_1 \times 3}$ by greedily maximizing mutual information with $\boldsymbol{\theta_{z_0}^*}$ using the algorithm proposed in [20] (Line 4 in Algorithm 1, $m_1$ can be a small number s.t. $m_1 << n$). The greedy algorithm exploits submodularity of the mutual information criterion, providing approximation guarantee of

$(1 - \frac{1}{e})$ of the optimum, with computational cost $\mathcal{O}(mn)$ for selection of $m$ out of $n$ locations.

The next step involves learning the local hyper-parameters $\mathbf{z}_{M_1}$ across the latent locations $X^*_{M_1}$ (see Line 5). The local hyper-parameters $\mathbf{z}_{M_1}$ are optimized jointly with global hyper-parameters $\boldsymbol{\theta}_{y_1}$ of the N-GP and hyper-parameters $\boldsymbol{\theta}_{z_1}$ of the stationary latent GP- $GP_z$. Joint optimization of the parameters set $\{\mathbf{z}_{M_1}, \boldsymbol{\theta}_{y_1}, \boldsymbol{\theta}_{z_1}\}$ can be done either by 1) maximizing *log* of Gaussian approximation in (8) of *marginal likelihood* $p(\mathbf{y_V}|X_V, X^*_{M_1}, \boldsymbol{\theta}_{y_1}, \boldsymbol{\theta}_{z_1}, \mathbf{z}_{M_1})$; or 2) Bayesian inference from Gaussian approximation in (9) of posterior $p(\boldsymbol{\theta}_{y_1}, \boldsymbol{\theta}_{z_1}, \mathbf{z}_{M_1}|X_V, \mathbf{y_V}, X^*_{M_1})$ using Monte Carlo [21] or Variational Methods [22], [23]. Hereafter, whenever we refer to joint optimization of local and global hyper-parameters using marginal likelihood, it is assumed that it can also be done using posterior sampling.

$$\{\boldsymbol{\theta}^*_{y_1}, \boldsymbol{\theta}^*_{z_1}, \mathbf{z}^*_{M_1}\} = argmax_{\boldsymbol{\theta}_{y_1}, \boldsymbol{\theta}_{z_1}, \mathbf{z}_{M_1}} p(\mathbf{y_V}|X_V, \boldsymbol{\theta}_{y_1}, \mathbf{z}^m_{V_1})$$
$$. p(\mathbf{z}^m_{V_1}|X_V, X^*_{M_1}, \mathbf{z}_{M_1}, \boldsymbol{\theta}_{z_1}) \quad (8)$$

$$\{\boldsymbol{\theta}^*_{y_1}, \boldsymbol{\theta}^*_{z_1}, \mathbf{z}^*_{M_1}\} = argmax_{\boldsymbol{\theta}_{y_1}, \boldsymbol{\theta}_{z_1}, \mathbf{z}_{M_1}} p(\mathbf{z}_{M_1}, \boldsymbol{\theta}_{z_1}|X_V,$$
$$X^*_{M_1}, \mathbf{z}^m_{V_1}).p(\boldsymbol{\theta}_{y_1}, \mathbf{z}^m_{V_1}|X_V, \mathbf{y_V}) \quad (9)$$

wherein $\mathbf{z}^m_{V_1}$ is predictive mean for local hyper-parameters across $X_V$, inferred by conditioning $\boldsymbol{\theta}_{z_1}$ on $\mathbf{z}_{M_1}$. $\mathbf{z}^m_{V_1}$ corresponds to, although suboptimally, latent space dynamics of the environmental phenomenon.

Since $\boldsymbol{\theta}_{z_1}$ is learned on $\mathbf{z}^m_{V_1}$, it represents the latent space dynamics better than $\boldsymbol{\theta}_{z_0}$, learned using the real dynamics. Correspondingly, the next set of latent locations $X_{M_2}$, learned by greedily maximizing mutual information using $\boldsymbol{\theta}^*_{z_1}$, will be more informative than learning $X_{M_2}$ using $\boldsymbol{\theta}^*_{z_0}$. Since the informativeness of the latent locations directly corresponds to learning accuracy of hyper-parameters (see Section III-B), improved informative learning of $X_{M_2}$ will then help in accurate learning of $\mathbf{z}_{M_2}$.

This improved learning concept can then be extended to iteratively learn the informative latent locations in an adaptive manner for $c$ iterations (line 7 and 8 in Algorithm 1). At the $i^{th}$ iteration, mutual information is maximized greedily to learn the next set of latent locations $X_{M_{i+1}}$ using $\boldsymbol{\theta}^*_{z_i}$ that is conditioned upon latent locations $\{X^*_{M_1}, \cdots, X^*_{M_i}\}$, as shown in (10).

$$X^*_{M_{i+1}} = argmax_{X_{M_{i+1}}} \mathbb{I}(\{X_{M_{i+1}}, X^*_{M_1 \cdots i}\}; X_{V \setminus M_1 \cdots i+1}$$
$$|\boldsymbol{\theta}^*_{z_i}, X^*_{M_1 \cdots i}) \quad (10)$$

Thereafter, for sensing the latent space dynamics across $X^*_{M_{i+1}}$ (line 8 in Algorithm 1), joint optimization of hyper-parameters $\{\boldsymbol{\theta}_{y_{i+1}}, \boldsymbol{\theta}_{z_{i+1}}, \mathbf{z}_{M_{i+1}}\}$ is performed by maximizing *marginal likelihood*, as shown in (11).

$$argmax_{\boldsymbol{\theta}_{y_{i+1}}, \boldsymbol{\theta}_{z_{i+1}}, \mathbf{z}_{M_{i+1}}} p(\mathbf{y_V}|X_V, \boldsymbol{\theta}_{y_{i+1}}, \mathbf{z}^m_{V_{i+1}}).p(\mathbf{z}^m_{V_{i+1}}$$
$$|X_V, \boldsymbol{\theta}_{z_{i+1}}, \mathbf{z}_{M_{i+1}}, X^*_{M_1 \cdots i+1}, \mathbf{z}^*_{M_1 \cdots i}) \quad (11)$$

wherein $\mathbf{z}^m_{V_{i+1}}$ is predictive mean for local hyper-parameters across $X_V$, inferred by conditioning $\boldsymbol{\theta}_{z_{i+1}}$ on $\{\mathbf{z}^*_{M_1}, \cdots, \mathbf{z}^*_{M_i}, \mathbf{z}_{M_{i+1}}\}$. Note that the local hyper-parameters learned in previous iterations, $\mathbf{z}^*_{M_1}, \cdots, \mathbf{z}^*_{M_i}$, are fixed under the joint optimization of $\{\boldsymbol{\theta}_{y_{i+1}}, \boldsymbol{\theta}_{z_{i+1}}, \mathbf{z}_{M_{i+1}}\}$.

Eventually, using *LISAL* algorithm, after $c$ iterations of adaptive sensing of the latent space dynamics, local hyper-parameters $\mathbf{z}^*_M = \{\mathbf{z}^*_{M_1}, \cdots, \mathbf{z}^*_{M_{c+1}}\}$ across latent locations $X^*_M = \{X^*_{M_1}, \cdots, X^*_{M_{c+1}}\}$, and global hyper-parameters $\boldsymbol{\theta}^*_y = \boldsymbol{\theta}^*_{y_{c+1}}, \boldsymbol{\theta}^*_z = \boldsymbol{\theta}^*_{z_{c+1}}$ are learned. Predictive mean $\mathbf{z}^{m*}_V$ that is inferred by conditioning $\boldsymbol{\theta}^*_z$ on $\{X^*_M, \mathbf{z}^*_M\}$, corresponds to the latent space dynamics across $X_V$.

Computational cost for learning latent locations $X^*_{M_1 \cdots c+1} \in \mathbb{R}^{m \times 3}$ using greedy maximization of mutual information is $\mathcal{O}(mn)$. Considering super polynomial computational cost, $\mathcal{O}(m^a)$, for optimizing an objective function in m-dimensional space, computational cost for learning the local hyper-parameters $\mathbf{z}^*_{M_1 \cdots c+1} \in \mathbb{R}^m$ across $m$ latent locations, under *LISAL*, is $\mathcal{O}(m_1^a + m_2^a)$. For small values of $m_1$ and $m_2$, computational cost under *LISAL* is significantly reduced as compared to the offline framework ($\mathcal{O}(m^a)$) proposed in [13], [14], [24].

## V. EMPIRICAL EVALUATION

We evaluate *LISAL* algorithm, using *LEIS* and *PCLSK* N-GP models, with three real world sensing datasets:

*1) Ozone Dataset:* consists of samples from year 1995 to 2011 for 60 stations across USA [2]. A sample for each station represents averaged ozone concentration for the whole year (see the dynamics in Figure 1(a)). 30 out of 60 locations were selected uniformly for training and the remaining 30 locations were used for testing (i.e. 480 samples for training and testing each).

*2) Wind Dataset:* consists of daily average wind speed (in knots = 0.5418 m/s) samples collected for year 1961 at 12 meteorological stations in the Republic of Ireland [1] (see the dynamics in Figure 1(b)). Samples from day 1 to day 351 (every 10 days) of 1961 for all 12 stations were used for training, and from day 5 to day 355 (every 10 days) were used for testing (i.e. 432 samples for training and testing each).

*3) Sunlight Dataset:* consists of uniformly separated 70 images, from 8:40 AM to 08:20 PM, collected to capture the light distribution under a forest canopy in San Jacinto mountains reserve [25]. Light intensity data for uniformly spread 180 spatial locations across the 70 timesteps was then simulated by averaging the pixel intensities (from the collected images) in the local grid around the selected image pixels (see the dynamics in Figure 1(c)). Training and test data constitutes light intensity samples across disjoint (though representative) sets of 36 (uniformly spread) spatial locations. Light Intensity samples from the first 12 images for the selected 36 spatial locations (constituting the dataset of size 432) were used for empirical evaluation. For evaluation of *LISAL* with varying number of training samples, light intensity datasets of varying sizes, from 216 to 1728, were prepared by varying the number of timesteps (i.e. images) from 6 to 48 for the selected 36 locations.

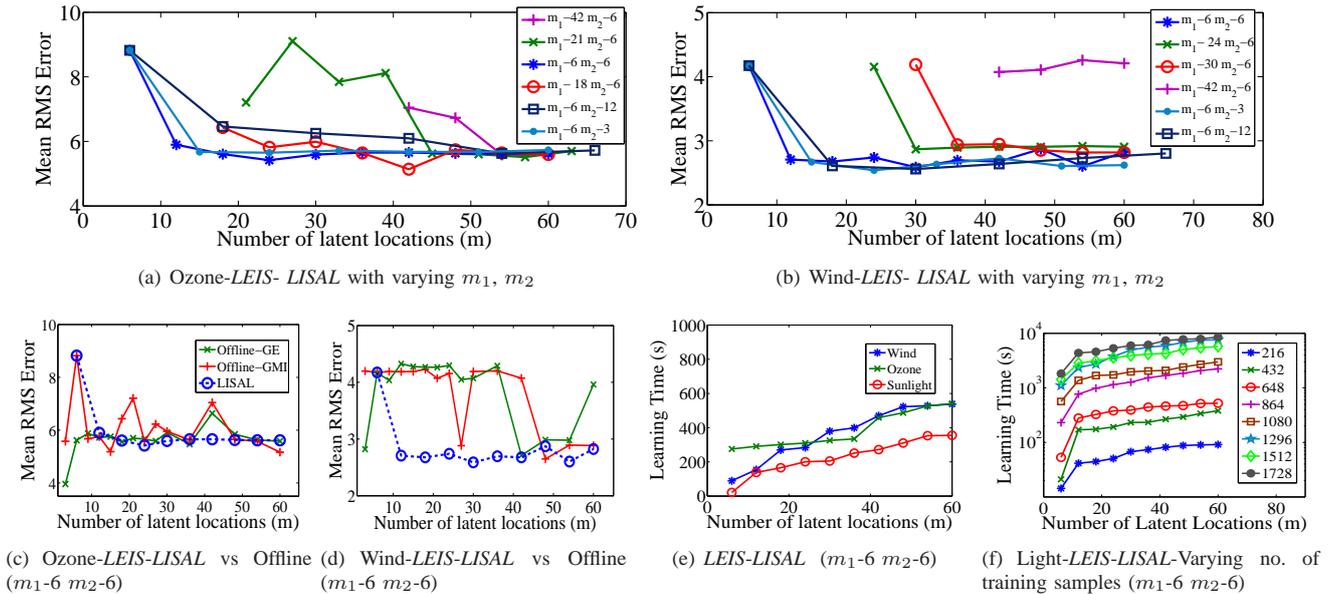

Fig. 2. Mean RMS error results for *LEIS* under *LISAL*

### A. Experimental Setup

We simulated a mobile robot path planning application, wherein the robot learned the next most informative location, by maximizing *mutual information*, using the N-GP considered, for sensing the real observation across the spatial expanse at a given timestep (a year for ozone data and a day for wind data, represented a timestep). Considering non-separable spatio-temporal covariance to be a more accurate choice for modeling environmental dynamics, examples 1 and 3 in [26] were used as covariance functions for deriving each of the N-GPs.

To simulate time delay for path traversal, the robot was constrained to make 6 observations at every timestep. These observations, together with the underlying GP model were used to predict the real phenomenon for all the test locations at a given timestep. Root Mean Square (RMS) error, between the predicted and the actual phenomenon value, was used to compare the accuracy of the underlying N-GP models.

### B. Empirical Results

We first evaluated the performance of *LISAL* algorithm for varying values of parameters $m_1$, $m_2$ and $c$, (as used in Algorithm 1 and discussed in Section IV), using the ozone and the wind training datasets. After each iteration of learning in Algorithm 1, starting from iteration 0 (line 5 in Algorithm 1), the learned *LEIS* model was evaluated within the experimental setup and the mean RMS error for the test run was calculated. Note that the $0^{th}$ iteration corresponds to the offline approach proposed in [14]. Referring to Figure 2(a) and 2(b), number of latent locations ($m$) is calculated as $m_1 + c * m_2$. High mean RMS value for the offline approach (i.e. iteration 0), was observed. However, in a few iterations of adaptive learning with *LISAL* (increasing the number of latent locations $m$), mean RMS error significantly decreased, and eventually converged. From analysis of the mean RMS plots for *LISAL* with varying values of $m_1$, $m_2$ in Figure 2(a), 2(b), we observe that *LISAL* performs well even for small values of $m_1$ and $m_2$. As a result, we fixed both $m_1$ and $m_2$ to a small value of 6 for the remaining experiments.

*LEIS* model with the offline approach for latent space dynamics (as proposed in [14]) was then learned using *entropy* and *mutual information* criteria. These learned models were then compared with *LEIS* model learned with (adaptive) *LISAL* algorithm. Comparison of the three *LEIS* models for Ozone and Wind datasets is presented in Figure 2(c) and 2(d) respectively. For the $0^{th}$ iteration (with $m = 6$), since *LISAL* is similar to the offline approach with mutual information, corresponding mean RMS error values were same. With increase in $m$, mean RMS error for the offline approach fluctuated whereas, mean RMS error for *LISAL* decreased in the initial few iterations (with adaptive increment in $m$), and then converged to an (possibly local) optima.

Figure 2(e) illustrates the computational time[1] for learning the *LEIS* model under *LISAL* for all the three datasets (with training samples ranging from 432 to 480). We observe that while the learning time increased linearly with increase in number of latent locations, overall time taken was small even for a large number of latent locations. Figure 2(f) evaluates the corresponding learning time variation with varying number of training samples size (from 216 to 1728), using the sunlight dataset. As expected, learning time increased polynomially with increase in the number of training samples (learning time for a GP is of the order of $\mathcal{O}(n^3)$ for $n$ training samples). However, increase in

---
[1]Experiments done using an i3 2.1 GHz CPU with 4 GB RAM

TABLE I
MEAN RMS ERROR COMPARISON FOR *LEIS*, *PCLSK*, *MGPs*, AND CORRESPONDING STATIONARY GPS.

(a) Sunlight-Ex.3

| Samples | S | *LEIS* |
|---|---|---|
| 216 | 45.62 | 45.33 |
| 432 | 49.35 | 45.37 |
| 648 | 46.77 | 44.31 |
| 864 | 51.85 | 46.90 |
| 1080 | 48.46 | 46.25 |
| 1296 | 47.68 | 46.39 |
| 1512 | 45.31 | 41.81 |
| 1728 | 44.32 | 42.34 |

(b) All Data

| GP | Ozone | Wind | Sunlight |
|---|---|---|---|
| S-Ex.3 | 5.65 | 2.98 | 49.30 |
| *LEIS*-Ex.3 | 5.08 | 2.86 | 45.31 |
| *PCLSK*-Ex.3 | 5.31 | 2.70 | 43.32 |
| *M2GPs*-Ex.3 | 5.72 | 3.02 | 47.01 |
| *M3GPs*-Ex.3 | 5.67 | 2.93 | 45.38 |
| *M5GPs*-Ex.3 | 5.81 | 2.93 | 45.38 |
| S-Ex.1 | 5.73 | 4.04 | 49.51 |
| *LEIS*-Ex.1 | 4.98 | 2.86 | 48.83 |

learning time with increment on $m$, was almost linear for different sizes of training samples. Note that, to show the efficiency of *LISAL* on a simple optimization setup, the learning optimizations, under *LISAL*, were performed using the standard *log marginal likelihood* (*lml*) maximization with ML-II approximation (page 109 in [4]). Learning time, using *LISAL*, will be further reduced by optimizing with either 1) *lml* maximization using Laplace Approximations or Bayesian Quadrature [27]; or 2) posterior sampling using Variational or Monte Carlo Methods.

Extensive experiments were also performed to compare multiple N-GP models (*LEIS*, *PCLSK* and Mixture of GPs (*MGPs*) [11], termed with prefix LEIS, PCLSK and MGPs in Table I) learned with the *LISAL* algorithm (with $c = 9$) and corresponding stationary models (S prefix, Ex.1 and Ex.3 from [26], termed accordingly as suffix in Table I) from which we derived the N-GP models. Mean RMS error comparison for all these models is presented in Table I. With fixed number of latent locations ($m = 60$), LEIS model learned with *LISAL*, performed consistently with increasing sample size (see Table I(a)). *LEIS* and *PCLSK*, learned using *LISAL*, resulted in mean RMS error improvement of up to 15% over the corresponding stationary GP models (see Table I(b)). We observe that MGPs with *LISAL* do not perform as consistently as the other two N-GP models - *LEIS* and *PCLSK*.

## VI. CONCLUSION AND FUTURE WORK

Proposed *LISAL* algorithm is an efficient approach for learning non-stationary dynamics associated with complex environmental phenomena, using Gaussian Process modeling. Extensive empirical evaluation with real world sensing datasets show that adaptively sensing the latent space dynamics, under *LISAL*, results in improved model learning (demonstrated using corresponding reduction in the RMS error) with very little overhead in the computation cost. Experiments with the complex spatio-temporal sunlight dataset of sample size up to 1700 demonstrate the scalability and applicability of *LISAL* for environmental applications. In the future, we plan to further improve upon the computation cost to efficiently learn N-GPs on very large datasets.